\documentclass[a4paper,twoside]{article}

\usepackage{epsfig}
\usepackage{subcaption}
\usepackage{calc}
\usepackage{amssymb}
\usepackage{amstext}
\usepackage{amsmath}
\usepackage{amsthm}
\usepackage{multicol}
\usepackage{pslatex}
\usepackage{apalike}
\usepackage{algorithm2e}
\usepackage[bottom]{footmisc}

\usepackage{multirow}
\usepackage{booktabs}
\usepackage{subcaption}
\usepackage{xcolor}
\usepackage[nolist]{acronym}
\usepackage{tikz}

\usepackage{listofitems}
\usetikzlibrary{arrows.meta}
\usetikzlibrary{positioning}
\colorlet{myred}{red!80!black}
\colorlet{myblue}{blue!80!black}
\colorlet{mygreen}{green!60!black}
\tikzset{
  >=latex, 
  node/.style={thick,circle,draw=myblue,minimum size=4,inner sep=0.5,outer sep=0.6},
  node in/.style={node,green!20!black,draw=mygreen!30!black,fill=mygreen!25},
  node hidden/.style={node,blue!20!black,draw=myblue!30!black,fill=myblue!20},
  node out/.style={node,red!20!black,draw=myred!30!black,fill=myred!20},
  connect arrow/.style={-{Latex[length=2.5,width=1.7]},ultra thin,black!80},
  node 1/.style={node in}, 
  node 2/.style={node hidden},
  node 3/.style={node out}
}
\def\nstyle{int(\lay<\Nnodlen?min(2,\lay):3)} 
\tikzset{
  grid/.style={draw=gray,thin},
  pics/mlp/.style={
    code={
      \readlist\Nnod{4,5,5,3} 

      \foreachitem \N \in \Nnod{ 
        \def\lay{\Ncnt} 
        \pgfmathsetmacro\prev{int(\Ncnt-1)} 

        \foreach \i [evaluate={\y=\N/2-\i; \x=\lay; \n=\nstyle;}] in {1,...,\N}{ 

          \node[node \n] (N\lay-\i) at (\x,\y) {};

          \ifnum\lay>1 
          \foreach \j in {1,...,\Nnod[\prev]}{ 
            \draw[connect arrow] (N\prev-\j) -- (N\lay-\i);
          }
          \fi 
        }
      }
    }
  }
}

\usepackage{siunitx}
\AtBeginDocument{\DeclareSIUnit{\kWp}{kWp}}

\usepackage{SCITEPRESS}     

\begin{document}

\title{Two-Stage Photovoltaic Forecasting: Separating Weather Prediction from Plant-Characteristics}

\author{\authorname{Philipp Danner\sup{1}\orcidAuthor{0000-0002-3005-630X} and Hermann de Meer\sup{1}\orcidAuthor{0000-0002-3466-8135}}
  \affiliation{\sup{1}University of Passau, Innstraße 41, Passau, Germany}
  \email{\{philipp.danner, hermann.demeer\}@uni-passau.de}
}

\keywords{
  Forecast Error,
  Photovoltaic System,
  Numerical Weather Prediction,
  Site-Specific Model
}

\abstract{
  Several energy management applications rely on accurate photovoltaic generation forecasts. Common metrics like mean absolute error or root-mean-square error, omit error-distribution details needed for stochastic optimization. In addition, several approaches use weather forecasts as inputs without analyzing the source of the prediction error. To overcome this gap, we decompose forecasting into a weather forecast model for environmental parameters such as solar irradiance and temperature and a plant characteristic model that captures site-specific parameters like panel orientation, temperature influence, or regular shading. Satellite-based weather observation serves as an intermediate layer.
  We analyze the error distribution of the high-resolution rapid-refresh numerical weather prediction model that covers the United States as a black-box model for weather forecasting and train an ensemble of neural networks on historical power output data for the plant characteristic model. Results show mean absolute error increases by 11\% and 68\% for two selected photovoltaic systems when using weather forecasts instead of satellite-based ground-truth weather observations as a perfect forecast. The generalized hyperbolic and Student's t distributions adequately fit the forecast errors across lead times.
}

\onecolumn{} \maketitle{} \normalsize{} \setcounter{footnote}{0} \vfill{}

\begin{acronym}
    \acro{PV}{Photovoltaic}
    
    \acro{MBE}{Mean Bias Error}
    \acro{MAE}{Mean Absolute Error}
    \acro{RMSE}{Root Means Square Error}
    \acro{MAPE}{Mean Absolute Percentage Error}

    \acro{MLP}{Multi-layer Perceptron}
    \acro{NWP}{Numerical Weather Prediction}
    
    \acro{DSWRF}{Downward Shortwave Radiation Flux}
    \acro{GHI}{Global Horizontal Irradiance}
    \acro{DNI}{Direct Normal Irradiance}
    \acro{DHI}{Diffuse Horizontal Irradiance}
    
    \acro{PVDAQ}{Photovoltaic Data Acquisition Public Datasets}
    \acro{HRRR}{High-Resolution Rapid Refresh}
\end{acronym}

\section{INTRODUCTION}
Many energy-management applications depend critically on accurate day-ahead forecasts of supply and demand. Examples include day-ahead electricity market bidding, where forecast errors translate directly to imbalance penalties~\cite{YANG2023108557}; microgrid operation and islanding strategies that must reserve storage capacity under uncertainty~\cite{Danner2024localres}; and building energy management that rely on reliable predictions to schedule flexibility and maximize self-consumption~\cite{danner2021}. In all these settings, the probabilistic structure of forecast errors, not only point-wise accuracy, drives risk-aware decisions.

Forecasts for weather-dependent energy resources such as \ac{PV} generation thus require realistic error models. Prior work often assumes Gaussian errors~\cite{Hemmati2020,Antoniadou-Plytaria2022} or adopts robust~\cite{Liu2017} and distributionally robust formulations~\cite{Shi2019} that use only partial distributional information. While these simplifications facilitate tractable optimization, they may mischaracterize empirical error structure or induce conservatism~\cite{roald_security_2015}. Moreover, common claims--e.g., that forecast variance monotonically increases with horizon~\cite{Antoniadou-Plytaria2022}--are not universally validated. Standard scalar metrics such as \ac{MAE} or \ac{RMSE} summarize overall accuracy but discard information about skewness, tail behavior, and temporal dependence that are essential for stochastic decision-making.

These gaps are addressed by (i) decomposing the forecasting chain into two interpretable components--a weather forecast module (capturing irradiance and temperature uncertainty) and a plant characteristic module (mapping weather to site-specific power, including orientation, shading, and temperature effects)--and (ii) empirically characterizing the lead-dependent error distributions and temporal correlations relevant for day-ahead planning.
Our analysis uses \ac{HRRR} \ac{NWP}, satellite-derived observations as an intermediate ground truth, and an ensemble of \acp{MLP} trained on multi-year \ac{PV} generation data. We quantify the error origin from meteorology and plant modelling, test common parametric families (normal, Student's t, generalized hyperbolic) against observed errors, and assess lead-wise temporal autocorrelation. Results provide empirically grounded guidance for selecting stochastic models and for post-processing strategies that reduce bias and tail risk in \ac{PV} forecasts.

The remainder of the paper reviews related work (Section~\ref{sec:related_work}), presents the two-part forecasting framework (Section~\ref{sec:methodology}), describes datasets and experimental design (Section~\ref{sec:experiment_setup}), reports the numerical analysis (Section~\ref{sec:discussion}), and concludes with implications for stochastic optimization (Section~\ref{sec:conclusions}).
\section{RELATED WORK}\label{sec:related_work}
\ac{PV} power forecasting is often classified by forecasting horizon~\cite{das2018forecasting}: very short-term (seconds-minutes) for smoothing and dispatch, short-term (hours-days) for operational planning and grid security, medium-term (weeks-month) for maintenance/scheduling, and long-term (months-years) for strategic planning and economics. This work focuses on day-ahead (short-term) forecasting.

Beyond classical statistical approaches, such as variants of auto regression models (applicable across all time scales), a range of weather-forecasting techniques is used for \ac{PV} power forecasting. For very short horizons, cloud-motion estimation from sky cameras is common; for short-term horizons, satellite imagery and \ac{NWP} models are widely used~\cite{Antonanzas2016}. Because raw weather forecasts are generally insufficient for \ac{PV} power forecasting, hybrid models that combine machine-learning components with weather predictions are common.

Typical machine-learning models include feed-forward neural networks
~\cite{markovics2022comparison,liu2015improved}, recurrent architectures such as Elman networks~\cite{chupong2011forecasting} and long short-term memory networks~\cite{danner2021,abdel2019accurate}, random forest regressors~\cite{almeida2015pv}, and ensemble learning methods~\cite{rana2015forecasting}. Common input features are recent power measurements and various weather parameters~\cite{markovics2022comparison}. Often \ac{NWP} variables such as irradiance, temperature, humidity, wind, cloud cover and pressure are used as weather input to the machine-learned models.
Other work search historical records for analog days~\cite{ding2011ann,liu2015improved}. These approaches fold weather-forecast error into the overall error without separating it, whereas some authors train on measured on-site weather (a perfect forecast) to isolate the power-mapping model error~\cite{cococcioni201124}.

Performance evaluations typically report scalar metrics such as \ac{MAE}, \ac{MAPE}, and \ac{RMSE}~\cite{das2018forecasting}. These metrics enable model comparison but provide limited insight into the actual error distribution. Only a few researchers analyze error distributions more deeply.

For instance, \cite{de2014photovoltaic} present a short-term neural-network forecasting method for a single site across horizons of 1, 3, 6, 12, and 24 hours. They report error distributions with slight skewness and decreasing kurtosis as the horizon increases. Their inputs include recent power, air temperature, module temperature, and plane-of-array irradiance parameters that require detailed measurements and are not generally available for many systems. Because their model is trained on measured weather, it assumes a perfect weather forecast and thus only reflects the error of the plant-level model.

For very short-term horizons, \cite{chu2015short} augment statistical models (auto-regression moving average, k-nearest neighbors) with a deterministic cloud-tracking and shadow-estimation model using sky images. Across 5, 10, and 15 minute horizons, the cloud-tracking approach achieved the lowest \ac{RMSE} and produced error distributions with near-zero skewness, moderate kurtosis, and short tails. These results indicate that including explicit short-term weather information can substantially improve forecasts. However, such findings do not directly generalize to day-ahead forecasting, where weather variability is larger.

A regional aggregation approach in \cite{junior2014forecasting} uses support vector regression and principal component analysis to forecast for six different plants spread over the island of Hokkaido (Japan). The error distributions show slight negative bias. These models use historical power and national weather forecasts (temperature, cloudiness, humidity, and horizontal extraterrestrial irradiance). Hence, weather-forecast errors are embedded in the aggregate error and are not disentangled. Results for aggregated systems do not necessarily transfer to single-sites.

In summary, hybrid models combining weather forecasts with data-driven site-specific models are the most realistic setting. However, the relative contribution of weather-forecast error versus site-model error remains unclear in many studies. Thus, we propose to analyze weather-forecast and plant-characteristic errors separately and to perform a detailed examination of error distributions for stochastic optimization.
\section{METHODOLOGY}\label{sec:methodology}
The \ac{PV} power forecasting is executed at time \( t \) and yields an output vector \( \hat{\mathbf{p}} \in \mathbb{R}^{f} \), representing predicted power for the subsequent \( f \) discrete forecast time steps (leads). One time step corresponds to one hour in this work, although higher temporal resolution is possible and limited only by data availability.

To analyze forecasting error, we decompose the problem into a plant characteristic model and a weather forecast model, reflecting distinct error sources. While the methodology is applicable to other weather-dependent renewables (e.g., wind), the present study focuses on \ac{PV} systems. The two models are described below.

\subsection{Plant Characteristic Model}
The plant characteristic model captures local, system-specific influences on \ac{PV} power output, including panel type, dimensions, rated power, orientation (tilt and azimuth), surface albedo, persistent shading from vegetation or buildings, and the effects of meteorological variables such as solar irradiance, air temperature, and wind. Because accurate on-site weather measurements are frequently unavailable, we employ satellite-based weather observations as an intermediate layer, enabling application over large spatial domains.

Although physically-based, non-linear models can reproduce \ac{PV} output with high fidelity~\cite{Holmgren2018}, they require detailed and accurate system and environmental parameters. In large \ac{PV} databases, this metadata is frequently missing, erroneous, or set to default values~\cite{Killinger2018}, and is difficult to extract reliably~\cite{danner2021a}. Moreover, local environmental conditions to determine shading and microclimate are seldom fully observed. Data-driven models therefore exploit empirical relationships among multiple inputs and may incorporate historical power traces to implicitly capture hardly observable system characteristics.

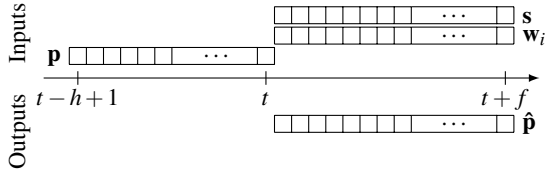
\begin{figure}[bt]
  \centering{}
  \begin{tikzpicture}[scale=0.9]
    \footnotesize{}

    \draw[->] (-3.25,0.0) -- (4,0.0);
    \draw[] (0,0.1) -- node[below, text height=0.25cm] {\( t \)} (0,-0.1);
    \draw[] (-2.75,0.1) -- node[below, text height=0.25cm] {\( t - h + 1 \)} (-2.75,-0.1);
    \draw[] (3.5,0.1) -- node[below, text height=0.25cm] {\( t + f \)} (3.5,-0.1);

    \draw[] (-3.6,0.1) node[right, rotate=90] {Inputs};
    \draw[] (-3.6,-0.1) node[left, rotate=90] {Outputs};

    \begin{scope}[xshift=-2.875cm, yshift=0.2cm]
      \draw[step=0.25cm] (0, 0) grid (1.5, 0.25);
      \draw[] (1.5,0) -- (2.75,0);
      \draw[] (2.175, 0.24) node[text height=0.25cm] {\dots{}};
      \draw[] (1.5,0.25) -- (2.75,0.25);
      \draw[step=0.25cm] (2.74, 0) grid (3.0, 0.25);
      \draw[] (0, 0.15) node[left, text height=0.25cm] {\( \mathbf{p} \)};
    \end{scope}
    \begin{scope}[xshift=0.125cm, yshift=0.5cm]
      \draw[step=0.25cm] (0, 0) grid (2.0, 0.25);
      \draw[] (2,0) -- (3.25,0);
      \draw[] (2.65, 0.24) node[text height=0.25cm] {\dots{}};
      \draw[] (2,0.25) -- (3.25,0.25);
      \draw[step=0.25cm] (3.24, 0) grid (3.5, 0.25);
      \draw[] (3.5, 0.15) node[right, text height=0.25cm] {\( \mathbf{w}_{i} \)};
    \end{scope}
    \begin{scope}[xshift=0.125cm, yshift=0.8cm]
      \draw[step=0.25cm] (0, 0) grid (2.0, 0.25);
      \draw[] (2,0) -- (3.25,0);
      \draw[] (2.65, 0.24) node[text height=0.25cm] {\dots{}};
      \draw[] (2,0.25) -- (3.25,0.25);
      \draw[step=0.25cm] (3.24, 0) grid (3.5, 0.25);
      \draw[] (3.5, 0.15) node[right, text height=0.25cm] {\( \mathbf{s} \)};
    \end{scope}

    \begin{scope}[xshift=0.125cm, yshift=-0.8cm]
      \draw[step=0.25cm] (0, 0) grid (2.0, 0.25);
      \draw[] (2,0) -- (3.25,0);
      \draw[] (2.65, 0.24) node[text height=0.25cm] {\dots{}};
      \draw[] (2,0.25) -- (3.25,0.25);
      \draw[step=0.25cm] (3.24, 0) grid (3.5, 0.25);
      \draw[] (3.5, 0.15) node[right, text height=0.25cm] {\( \mathbf{\hat{p}} \)};
    \end{scope}
  \end{tikzpicture}
  \caption{Inputs/outputs of the plant characteristic model on the time axis.}
  \label{fig:input_output_time}
\end{figure}

Following prior work, we include the historical power vector \( \mathbf{p}\in\mathbb{R}^{h} \), comprising the \( h \) observations from \( t-h+1 \) to \( t \). The lags share the same temporal resolution as the forecast horizon \( f \) (here hourly). In addition, we incorporate \( n \) weather features \( \mathbf{w}_{i}\in\mathbb{R}^{f},\; i \in \{ 1,\dots,n \},\; n\in\mathbb{N} \), selected by a correlation analysis to retain variables most informative for power generation.
During training, satellite-derived weather observations at the PV site are used as targets for the plant model rather than \ac{NWP} forecasts. This ensures the model learns site- and system-specific mappings (e.g., orientation, shading, albedo) independently of \ac{NWP} errors. Retrieval of satellite observations requires the PV system's geographic coordinates (latitude and longitude).
The solar elevation angles \( \mathbf{s}\in\mathbb{R}^{f} \) are computed following~\cite{hughes1985}. Atmospheric refraction is neglected at the hourly resolution considered. Angles below zero are set to zero to indicate nighttime; the elevation attains a maximum of \( 90^\circ \) at the zenith.
All inputs and outputs are summarized in Figure~\ref{fig:input_output_time}.

\begin{figure}[tb]
  \centering{}
  \begin{tikzpicture}[x=10cm,y=6cm,scale=0.85]
    \footnotesize{}

    \draw[->] (0.2,0.4) -- node[right, text height=0.1cm] {\( \mathbf{p} \)} (0.2,0.3);
    \draw[->] (0.3,0.4) -- node[right, text height=0.1cm] {\( \mathbf{w}_{i} \)} (0.3,0.3);
    \draw[->] (0.4,0.4) -- node[right, text height=0.1cm] {\( \mathbf{s} \)} (0.4,0.3);

    \draw[] (0.0,0.3) -- (0.6,0.3);

    \draw[] (0.0,0) node[draw,label={[above left]:\( {MLP}_{1} \)}] (mlp_1){
      \begin{tikzpicture}
        \draw[] pic[x=0.5cm,y=0.4cm,rotate=270] {mlp};
      \end{tikzpicture}
    };
    \draw[] (0.25,0) node[draw,label={[above left]:\( {MLP}_{2} \)}] (mlp_2){
      \begin{tikzpicture}
        \draw[] pic[x=0.5cm,y=0.4cm,rotate=270] {mlp};
      \end{tikzpicture}
    };
    \draw[] (0.425,0) node {\dots{}};
    \draw[] (0.6,0) node[draw,label={[above left]:\( {MLP}_{m} \)}] (mlp_m){
      \begin{tikzpicture}
        \draw[] pic[x=0.5cm,y=0.4cm,rotate=270] {mlp};
      \end{tikzpicture}
    };

    \draw[->] (0.0,0.3) -- (mlp_1);
    \draw[->] (0.25,0.3) -- (mlp_2);
    \draw[->] (0.6,0.3) -- (mlp_m);

    \draw[] (0.3,-0.332) node[draw, minimum width=7cm] (combined){Mean output vector};

    \draw[->] (mlp_1) -- node[right] {\( \mathbf{\hat{p}_1} \)} (mlp_1 |- (0,-0.285););
    \draw[->] (mlp_2) -- node[right] {\( \mathbf{\hat{p}_2} \)} (mlp_2 |- (0,-0.285););
    \draw[->] (mlp_m) -- node[right] {\( \mathbf{\hat{p}_m} \)} (mlp_m |- (0,-0.285););
    \draw[->] (0.3,-0.38) -- node[right] {\( \mathbf{\hat{p}} \)} (0.3,-0.465);

  \end{tikzpicture}
  \caption{Plant characteristic model: ensemble of \acp{MLP}.}
  \label{fig:ensemble_NN}
\end{figure}
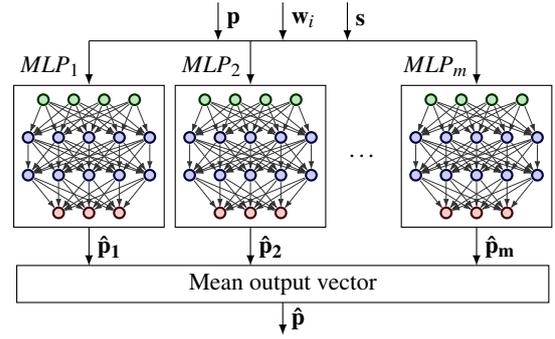

The plant characteristic model is implemented as a \ac{MLP}, a feed-forward network with one input layer, one or more hidden layers, and a linear output layer. Neurons are fully connected between consecutive layers; each neuron applies a weighted sum followed by a non-linear activation. Activation functions and other hyperparameters are selected by tuning. Input features \( \mathbf{p} \), \( \mathbf{w}_{i} \), and \( \mathbf{s} \) are scaled using a min-max scaler fitted on the training set; the same factors are applied to test data.
\ac{MLP} training minimizes the mean squared error using gradient-based optimization. Given the non-convexity of the loss surface and sensitivity to initialization, we train an ensemble of \( m \) instances with different random initialization. The element-wise mean of the individual output vectors \(\mathbf{\hat{p}}_{j}\in\mathbb{R}^{f}\), \( j \in \{1,\dots,m\} \), constitutes the final prediction (see Figure~\ref{fig:ensemble_NN}). This bagging approach~\cite{Breiman1996} reduces variance while preserving the temporal structure of inputs across ensemble members.

\subsection{Weather Forecast Model}
The weather forecast model provides predictions of meteorological variables that drive \ac{PV} power generation, such as surface irradiance and air temperature, which in turn reflect larger-scale atmospheric processes (clouds, advection, diurnal rotation, etc.).

While pure machine-learning weather prediction models, like GraphCast~\cite{Lam2023}, have shown promise, \ac{NWP} still remains the operational standard for short-term forecasting~\cite{Antonanzas2016,junior2014forecasting,almeida2015pv}. \ac{NWP} integrates the governing equations of atmospheric dynamics and thermodynamics, initialized from an assimilated state composed of heterogeneous observations (satellites, surface stations, radars, radiosondes), and produces prognostic fields (temperature, humidity, wind, pressure, radiative fluxes, etc.) at prescribed spatial and temporal resolutions. High-resolution \ac{NWP} systems, such as the \ac{HRRR} system, require substantial computational resources and therefore employ parameterization for sub-grid processes (cloud microphysics, convection, turbulence, radiation), which introduce structural approximations~\cite{hrrr2022a}.

Given the complexity and computational cost of \ac{NWP}, we treat it as a black box and focus on quantifying its forecast errors for meteorological fields relevant to \ac{PV} generation. Specifically, we assess \ac{NWP} errors by comparing forecasted meteorological variables with satellite-derived observations, the latter serving as an intermediate representation between \ac{NWP} output and the plant characteristic model.

\section{EXPERIMENTAL SETUP}\label{sec:experiment_setup}
In the following, the dataset and model parameterization are detailed. Further, the extraction procedure for the error distribution is explained.

\subsection{Plant Characteristic Model}

\subsubsection{Dataset}
\begin{figure}[tb]
  \centering{}
  \includegraphics[width=\linewidth]{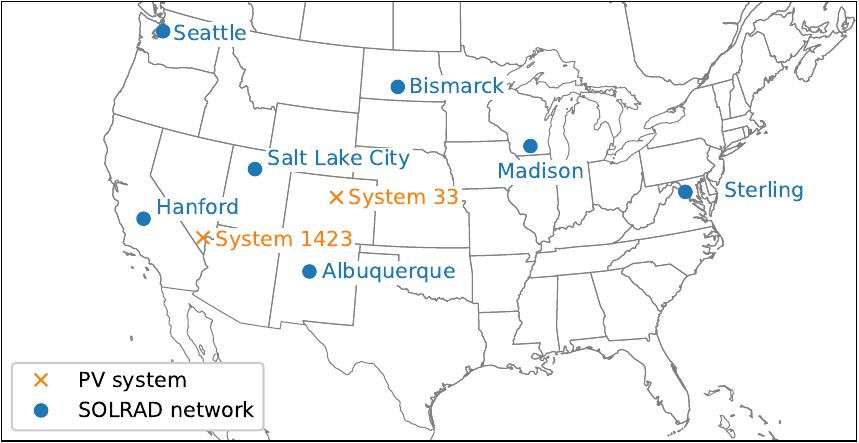}
  \caption{Location of selected \ac{PV} systems and the SOLRAD irradiance observation network.}
  \label{fig:map}
\end{figure}

Two \ac{PV} systems have been selected from the PVDAQ dataset~\cite{pvdaq2021}: system 33 (Colorado) and system 1423 (Nevada), shown in Figure~\ref{fig:map}, with rated peak powers of 2.4~kW and 6~kW, respectively. Given their modest rated peak power, generation curtailment due to grid constraints is not anticipated. The sites have been chosen for their continental locations and fixed panel orientations, making them broadly representative of the majority of PVDAQ installations, and because they are the only sites with sufficient recent measurements that overlap the temporal coverage of the employed \ac{NWP} dataset.

Raw inverter AC power measurements have been pre-processed by removing invalid records (extreme outliers) and by setting small negative values, attributable to standby losses during non-generation periods, to zero. Data with heterogeneous sampling intervals (1 to 15 min) have been aggregated to an hourly resolution by computing period means, matching the temporal granularity of available weather forecasts. Each system's power time series was normalized by its rated peak power.

\begin{figure}[tb]
  \centering{}
  \includegraphics[width=\linewidth]{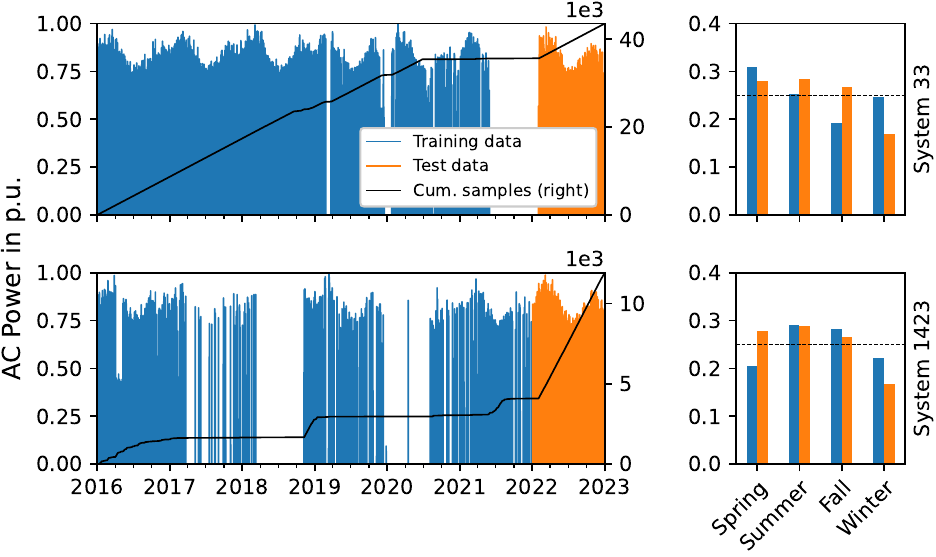}
  \caption{Pre-processed AC power data from the selected \ac{PV} systems with training and test data split.}
  \label{fig:ac_power}
\end{figure}

The training set comprises multi-year data from 2016-2021, while the test set comprises the calendar year 2022; both sets are illustrated in Figure~\ref{fig:ac_power}. Samples have been excluded from training and evaluation if they lacked at least \( h - 1 \) preceding valid observations or \( f \) subsequent valid leads. The cumulative-valid-samples plot (Figure~\ref{fig:ac_power}) indicates that system 33 provides approximately four years of valid training data (67.7\% of the training period) and 90.9\% coverage of the testing year; system 1423 provides under six months of valid training data and 89.5\% of the testing year. For both systems the data are well distributed across seasons and hours of day (minimum 16.7\% per season), supporting representative evaluation despite the smaller sample size for system 1423.

Historical satellite-derived weather observations from Solcast have been used to train and evaluate the plant characteristic model. Solcast provides data from 2007 up to seven days prior to the present, with native temporal resolution of 5~min, spatial resolution of 2~km for irradiance/cloud variables and 27~km for ancillary parameters such as air temperature. For each PV system the Solcast time series have been spatially interpolated to the site location and temporally resampled to hourly averages. Prior validation studies report low bias (0.33\%) and small standard deviation (1.7\%) for Solcast \ac{GHI}~\cite{Bright2019}; we reproduced these findings by comparison with SOLRAD ground measurements (Figure~\ref{fig:map}), supporting the use of Solcast irradiance as ground-truth for weather inputs. Candidate meteorological features for the weather data included \ac{GHI}, \ac{DHI}, \ac{DNI}, air temperature, and wind speed.

\subsubsection{Input Feature Selection}
For feature selection, a correlation analysis of historical power output and weather parameters with the power output in one time step has been conducted. Both Pearson and Spearman correlations of the \ac{PV} systems are listed in Table~\ref{tab:correlation}.

\begin{table}[bt]
  \caption{Pearson (P) and Spearman (S) correlation of input features to \ac{PV} power output.}
  \label{tab:correlation}
  \footnotesize{}
  \centering{}
  \begin{tabular}{ r c c c c c c }
    \toprule{}
    System     & \multicolumn{2}{c}{33} & \multicolumn{2}{c}{1423} & \multicolumn{2}{c}{both}  \\
    Corr.     & P      & S     & P    & S    & P    & S     \\
    \midrule{}
    GHI             & 0.91         & 0.88         & 0.93       & 0.97        & 0.91       & 0.89         \\
    \( s \) & 0.79         & 0.86         & 0.87       & 0.96        & 0.81       & 0.87         \\
    Temp.           & 0.40         & 0.54         & 0.23       & 0.29        & 0.34       & 0.34          \\
    \( \mathbf{p}_{t-1} \)           & 0.90         & 0.90         & 0.93       & 0.93        & 0.91       & 0.91         \\
    \( \mathbf{p}_{t-24} \)          & 0.80         & 0.89         & 0.93       & 0.98        & 0.82       & 0.92         \\
    \bottomrule{}
  \end{tabular}
\end{table}
From the analysis of previous-hour lags of power generation, the auto-correlation exhibits the largest average Pearson coefficients across both \ac{PV} systems with \( \mathbf{p}_{t-1} \) (0.91) and \( \mathbf{p}_{t-24} \) (0.82). Such strong persistence motivates the use of simple benchmarks like Same-as-Last and Same-as-Yesterday. Auto-correlations for intermediate lags (2-23 hours) and for lags exceeding one day are substantially lower. Since we forecast simultaneously for \( f = 24 \) leads, we therefore employ \( h = 24 \) hourly lags as input to ensure the one-day temporal dependence is available for each element of the output vector \(\mathbf{\hat{p}}\); the \ac{MLP} may additionally learn useful structure from other lags.

\begin{figure}[tb]
  \centering{}
  \includegraphics[width=0.8\linewidth]{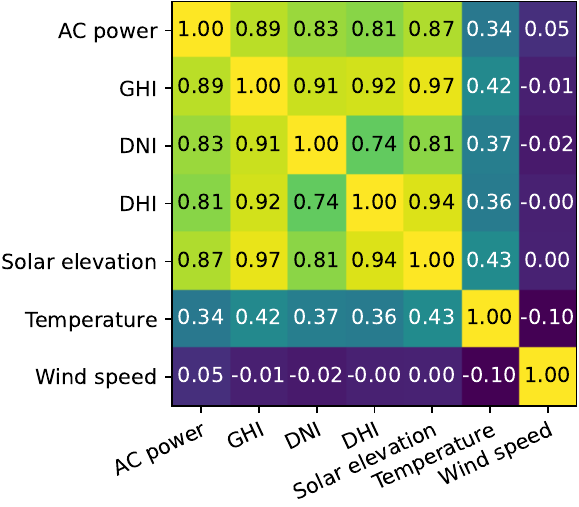}
  \caption{Spearman correlation of input features to the AC power output of both \ac{PV} systems.}
  \label{fig:input_feature_correlation}
\end{figure}

Among candidate meteorological features (\ac{GHI}, \ac{DNI}, \ac{DHI}, air temperature, wind speed), irradiance-related variables show the strongest association with power (Figure~\ref{fig:input_feature_correlation}). To avoid multicollinearity and reduce input dimensionality, we retain \ac{GHI} as the sole irradiance feature; its mean Pearson and Spearman correlations with power are 0.91 and 0.89, respectively (averaged over both sites).
Air temperature exhibits moderate association (Spearman 0.54 for system 33, 0.29 for system 1423) and is retained because it affects module efficiency, known from physical models~\cite{Holmgren2018}, and conveys seasonal information, with lower temperatures in winter and higher temperatures in summer in the Northern Hemisphere.
Although a cooling effect of wind is known from physical models, its impact on the power output of a \ac{PV} system is minor~\cite{zhu2015power}. As the Spearman correlation of wind speed to power output is negligible at 0.05, it is not used as input feature.

The solar elevation angle \( s \), calculated from location and time, attains a high Spearman correlation (0.87). Although correlated with \ac{GHI}, \( s \) encodes complementary geometric and temporal information (timing of sunrise/sunset and diurnal progression) that is not solely determined by irradiance, e.g., in case of clouds, and thus is included as an input feature.

\subsubsection{Hyperparameter Tuning}
We tuned the \ac{MLP} hyperparameters via an exhaustive grid search over a realistic parameter space. Each candidate configuration was assessed using five-fold cross-validation on the training set to obtain unbiased estimates of generalization performance and to mitigate overfitting. Architectural choices were guided by established best practices~\cite{heaton2008}: while a single hidden layer is sufficient to approximate non-linear mappings given adequate width, preliminary experiments showed no consistent benefit from architectures with more than two hidden layers. To avoid overfitting, the hidden-layer widths were constrained. A practical reference point is two-thirds of the number of input neurons plus the number of outputs (88 for our setup). Additionally, we imposed an upper bound of twice the input dimension (192) and required hidden-layer width to lie between the input (24) and output (96) dimensions.

\begin{figure}[tb]
  \centering{}
  \includegraphics[width=0.94\linewidth]{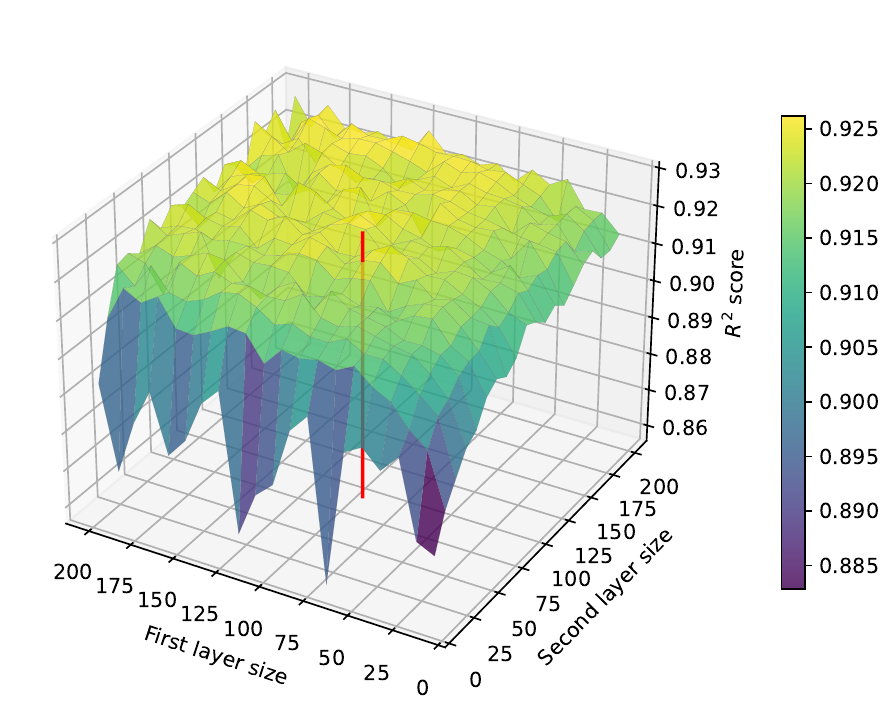}
  \caption{Impact of hidden layer width on the \( R^2 \) score for system 33. The red indicator shows the selected width, highlighting the start of the yellowish plateau.}
  \label{fig:hp_neurons_hidden_layer}
\end{figure}

Within the ensemble, all \acp{MLP} are identically configured with two hidden layers using rectified linear unit (ReLU) activation \( f(x) = max(0, x) \)~\cite{nair2010} (superior to identity, logistic and tanh), while the output-layer neurons use the identity activation. To determine the optimal width, we evaluated the cross-validated \( R^2 \) score for varying neuron counts per hidden layer. As illustrated for system 33 in Figure~\ref{fig:hp_neurons_hidden_layer}, performance increases between 10 and 50 neurons per layer and plateaus beyond around 90 neurons. Thus, 90 and 80 neurons are used on first and second hidden layers, respectively, balancing predictive performance and overfitting risk; system 1423 exhibited similar behavior. Finally, we set the ensemble size \( m = 200 \), reducing \ac{MAE} from 3.65\% (average of 30 individual \acp{MLP}) to 2.81\% while keeping training time below one minute on a 16-core machine.

\subsection{Weather Forecast Model}
The \ac{HRRR} weather forecast model provides high-resolution forecasts for the United States and is regarded state-of-the-art for short-term \ac{NWP}~\cite{hrrr2022a}. The current operational version (HRRRv4, deployed December 2\textsuperscript{nd}, 2020) produces hourly forecasts on a 3~km horizontal grid with an 18-hour horizon; forecasts are extended to 48~hours in cycles issued every 6~hours. HRRRv4 provides surface solar radiation among its prognostic variables and exhibits substantially improved irradiance skill relative to HRRRv3~\cite{hrrr2022b}.

Consistent with the plant characteristic feature analysis, we evaluate \ac{HRRR} forecasts of \ac{GHI} and air temperature at ground. For each PV site we selected the nearest HRRR grid point to represent the model forecast at the location. Solcast satellite-derived fields serve as the observational reference. To increase climatic representativeness, we also evaluate \ac{HRRR} forecasts at locations from the SOLRAD ground network shown in Figure~\ref{fig:map}.
Due to differing grid definitions and resolutions, the HRRR-Solcast mapping can introduce a spatial offset of up to approximately 1.41~km. This mapping error may contribute to the forecast-observation discrepancies. We suspect the error is small enough to be negligible as the weather on an hourly scale is typically a larger phenomenon.

\subsection{Error Analysis}
We analyze the performance of the plant characteristic and weather forecast model individually using \ac{MBE} in Equation~\eqref{eq:mbe} and \ac{MAE} in Equation~\eqref{eq:mae} where \( \tilde{p} \in \mathbb{R}^f \) is the ground-truth output between \( t + 1 \) and \( t + f \). Further, the combined model is compared with the individual models to get insights into the error source.

\begin{equation}\label{eq:mbe}
  MBE = \frac{1}{n} \sum{} \left(\frac{1}{f} \sum{} (\hat{p} - \tilde{p}) \right)
\end{equation}

\begin{equation}\label{eq:mae}
  MAE = \frac{1}{n} \sum{} \left(\frac{1}{f} \sum{} |\hat{p} - \tilde{p}| \right)
\end{equation}

Beyond aggregate performance metrics, we analyze empirical forecast-error distributions by lead time using the day-ahead forecast initialized at 06:00 UTC for each day; 06:00 UTC corresponds to 23:00 local standard time for system 33 and 22:00 for system 1423. We assume that seasonal variations in air temperature, \ac{GHI}, and hence \ac{PV} generation are captured by the weather-forecast model, so that the residual forecast errors are approximately independent of the seasonal cycle and may be characterized by a common lead-dependent distribution. This procedure is consistent with prior studies~\cite{ding2021distributionally,de2014photovoltaic}.

\section{RESULTS AND DISCUSSION}\label{sec:discussion}
We first analyze plant-model errors under perfect forecasts, then evaluate weather-forecast errors, and finally examine combined errors.

\subsection{Plant Characteristic Model}
The plant characteristic model has been evaluated without the influence of weather-forecast errors using Solcast satellite data as ground truth (perfect forecast conditions). The model has been applied for the year 2022 with forecasts initialized daily at 06:00~UTC and a forecasting horizon of 24 hourly leads. Errors are normalized by the system peak power (\SI[per-mode=symbol]{}{\watt\per\kWp}).

\begin{figure}[tb]
  \centering{}
  \begin{subfigure}{0.9\linewidth}
    \includegraphics[width=\linewidth]{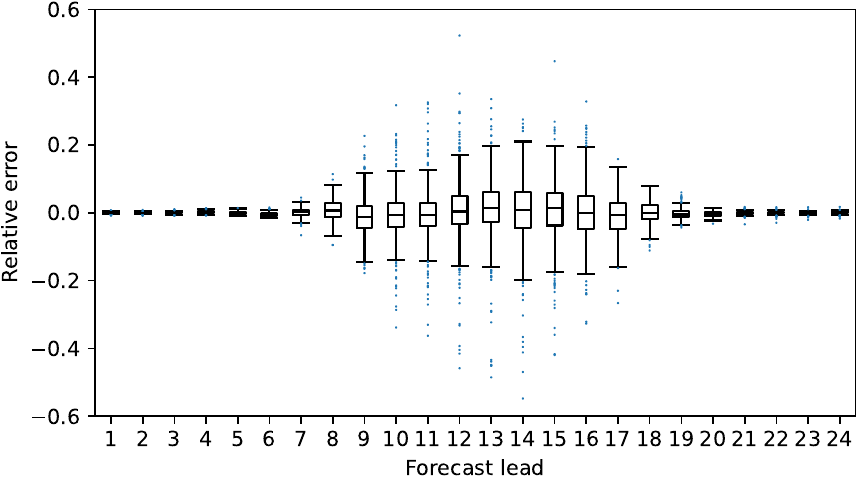}
    \caption{Relative error of different leads.}
    \label{fig:pcm_33_boxplot}
    \vspace{0.2cm}
  \end{subfigure}
  \begin{subfigure}{0.45\linewidth}
    \includegraphics[width=\linewidth]{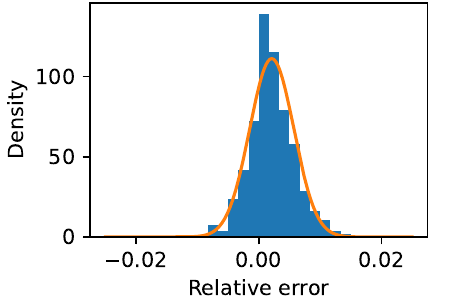}
    \caption{Distribution for lead 4}
    \label{fig:pcm_33_lead_4}
    \vspace{0.1cm}
  \end{subfigure}
  \begin{subfigure}{0.45\linewidth}
    \includegraphics[width=\linewidth]{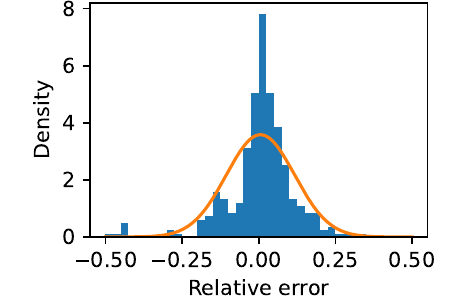}
    \caption{Distribution for lead 13}
    \label{fig:pcm_33_lead_13}
    \vspace{0.1cm}
  \end{subfigure}
  \caption{Errors of the plant characteristic model with a forecast at 23:00 of system 33. Orange curves show the fitted normal distribution in the probability density plots.}
\end{figure}

Both \ac{PV} systems exhibit a \ac{MAE} of approximately 2.8\% of peak power (\( \approx \) \SI[per-mode=symbol]{28}{\watt\per\kWp}). The \ac{MBE} is negligible for system~33 (\SI[per-mode=symbol]{-0.12}{\watt\per\kWp}) but shows a negative bias for system~1423 (\SI[per-mode=symbol]{-5.53}{\watt\per\kWp}), see Table~\ref{tab:error_summary}. The underestimation for system~1423 is most pronounced during daytime (leads~7-19) and may be attributable to its smaller training dataset.

\begin{table}[tb]
  \caption{Error metrics for 24 leads starting at 6:00~UTC.}
  \label{tab:error_summary}
  \footnotesize{}
  \centering{}
  \begin{tabular}{ c c c c c c c c c }
    \toprule{} %
    &        & Plant         & \multicolumn{2}{c}{Weather} & Combined                       \\ %
    \multicolumn{2}{r}{System} & Power         & GHI                         & Temp           & Power         \\ %
    &        & \SI[per-mode=symbol]{}{\watt\per\kWp} & \SI[per-mode=symbol]{}{\watt\per\metre\squared}              & \SI{}{\degreeCelsius} & \SI[per-mode=symbol]{}{\watt\per\kWp} \\
    \midrule%
    \multirow{2}{*}{\rotatebox[origin=c]{90}{MBE}} & 33     & -0.12         & 20.68                       & 3.88           & 18.34         \\ \vspace{0.1cm} %
    & 1423   & -5.53         & 12.66                       & -1.02          & 6.57          \\
    \multirow{2}{*}{\rotatebox[origin=c]{90}{MAE}} & 33     & 28.12         & 39.94                       & 3.97           & 47.28         \\ 
    & 1423   & 28.04         & 22.74                       & 1.61           & 31.16         \\
    \bottomrule{} %
  \end{tabular}
\end{table}

\begin{figure}[tb]
  \centering{}
  \begin{subfigure}{0.9\linewidth}
    \includegraphics[width=\linewidth]{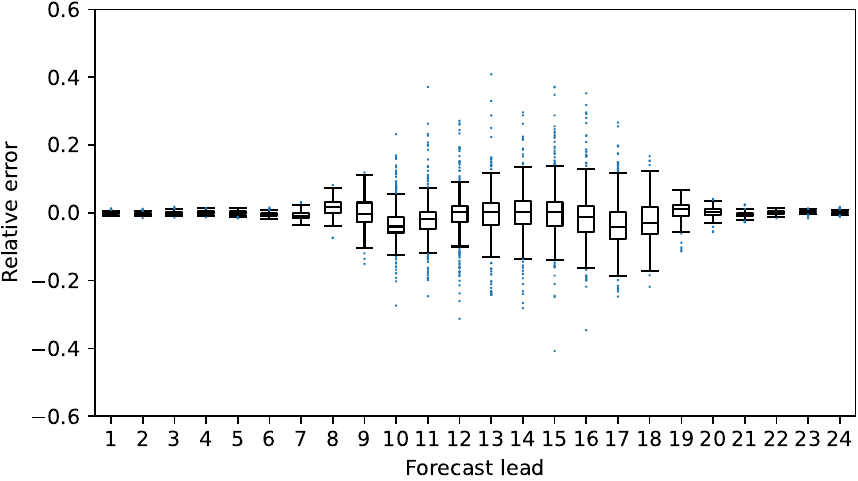}
    \caption{Relative error of different leads.}
    \label{fig:pcm_1423_boxplot}
    \vspace{0.2cm}
  \end{subfigure}
  \begin{subfigure}{0.45\linewidth}
    \includegraphics[width=\linewidth]{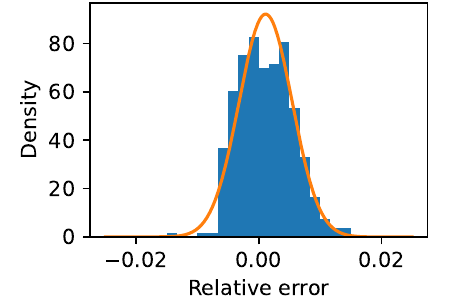}
    \caption{Distribution for lead 4}
    \label{fig:pcm_1423_lead_4}
    \vspace{0.1cm}
  \end{subfigure}
  \begin{subfigure}{0.45\linewidth}
    \includegraphics[width=\linewidth]{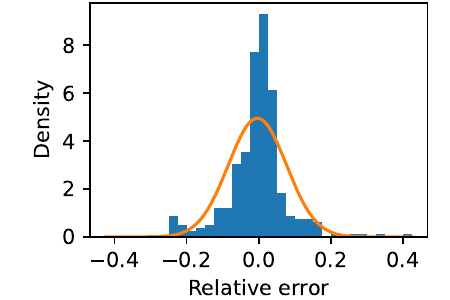}
    \caption{Distribution for lead 13}
    \label{fig:pcm_1423_lead_13}
    \vspace{0.1cm}
  \end{subfigure}
  \caption{Errors of the plant characteristic model with a forecast at 22:00 of system 1423. Orange curves show the fitted normal distribution in the probability density plots.}
\end{figure}

Error distributions are unimodal across all leads. Contrary to expectation, small nonzero errors persist during nocturnal leads (1-6 and 20-24). These nocturnal errors are model noise and are approximately normally distributed with low dispersion (cf. Figures~\ref{fig:pcm_33_lead_4} and~\ref{fig:pcm_1423_lead_4}). Applying a sunrise/sunset filter could avoid these nighttime errors.
During daytime, the error distributions are approximately symmetric (examples: Figures~\ref{fig:pcm_33_lead_13} and~\ref{fig:pcm_1423_lead_13}), with an average absolute skewness of about 0.5 (moderate). Error mass is concentrated near zero, indicating frequent accurate predictions, while occasional large outliers occur. This behavior is reflected in elevated kurtosis: mean kurtosis of 5.67 for system~33 and 4.87 for system~1423, indicating leptokurtic (peaked and heavy-tailed) distributions. The minimum kurtosis falls slightly below 3 for a few nocturnal leads.

\subsection{Weather Forecast Model}\label{sec:discussion_weather}
Irradiation has a higher impact on the power output of the \ac{PV} systems, thus we analyze the \ac{GHI} forecast error of the weather forecast model in detail. Temperature forecast errors are shown in the Appendix.

\begin{figure}[tb]
  \centering{}
  \begin{subfigure}{0.9\linewidth}
    \includegraphics[width=\linewidth]{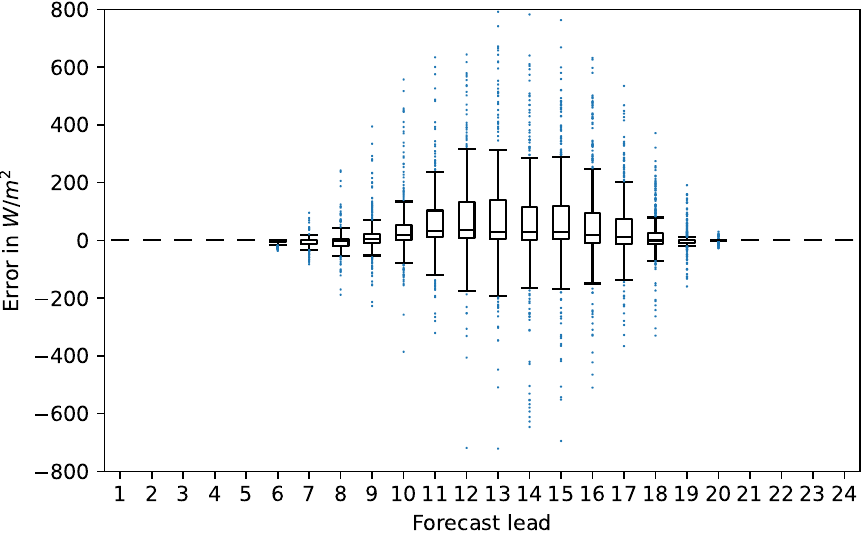}
    \caption{Relative error of different leads.}
    \label{fig:wfm_33_boxplot}
    \vspace{0.2cm}
  \end{subfigure}
  \begin{subfigure}{0.45\linewidth}
    \includegraphics[width=\linewidth]{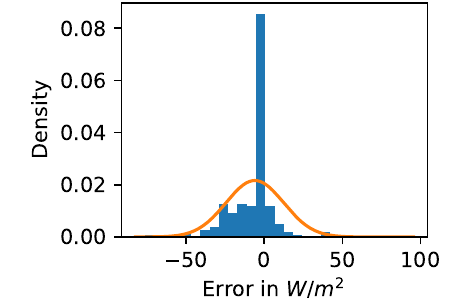}
    \caption{Distribution for lead 7}
    \label{fig:wfm_33_lead_7}
    \vspace{0.1cm}
  \end{subfigure}
  \begin{subfigure}{0.45\linewidth}
    \includegraphics[width=\linewidth]{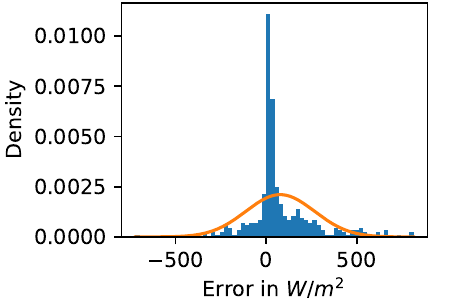}
    \caption{Distribution for lead 13}
    \label{fig:wfm_33_lead_13}
    \vspace{0.1cm}
  \end{subfigure}
  \caption{Errors of the weather forecast model with a forecast at 23:00 of system 33. Orange curves show the fitted normal distribution in the probability density plots.}
\end{figure}

\begin{figure}[tb]
  \centering{}
  \begin{subfigure}{0.9\linewidth}
    \includegraphics[width=\linewidth]{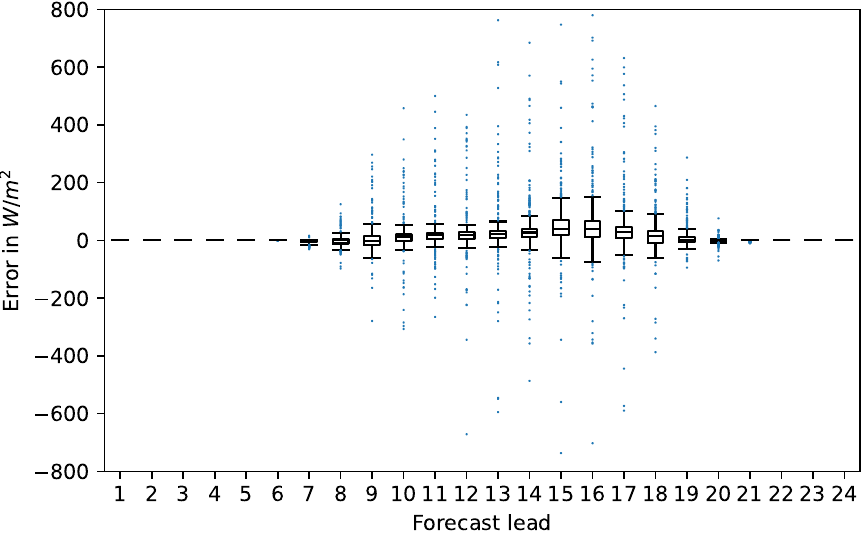}
    \caption{Relative error of different leads.}
    \label{fig:wfm_1423_boxplot}
    \vspace{0.2cm}
  \end{subfigure}
  \begin{subfigure}{0.45\linewidth}
    \includegraphics[width=\linewidth]{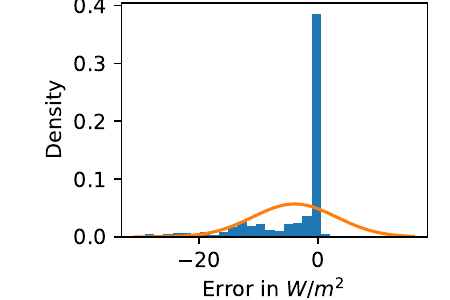}
    \caption{Distribution for lead 7}
    \label{fig:wfm_1423_lead_7}
    \vspace{0.1cm}
  \end{subfigure}
  \begin{subfigure}{0.45\linewidth}
    \includegraphics[width=\linewidth]{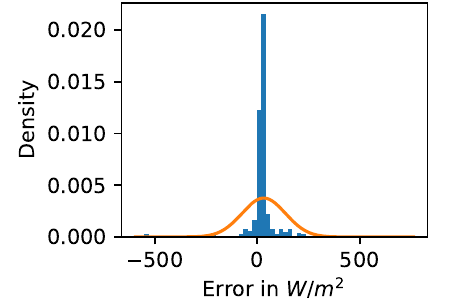}
    \caption{Distribution for lead 13}
    \label{fig:wfm_1423_lead_13}
    \vspace{0.1cm}
  \end{subfigure}
  \caption{Errors of the weather forecast model with a forecast at 23:00 of system 1423. Orange curves show the fitted normal distribution in the probability density plots.}
\end{figure}

\ac{GHI} forecast errors are near zero at night, as indicated for system~33 in Figure~\ref{fig:wfm_33_boxplot} and system~1423 in Figure~\ref{fig:wfm_1423_boxplot}. During daytime both locations show a systematic positive bias, which is also present across the SOLRAD stations. This systematic overestimation across stations is \SI[per-mode=symbol]{11.75}{\watt\per\metre\squared} (Table~\ref{tab:weather_forecast-all_stations}) and has been reported previously~\cite{hrrr2022b}.
Temperature forecasts exhibit a substantial positive bias at system~33 (\SI{3.88}{\degreeCelsius}), whereas the spatial average bias is small (\SI{0.30}{\degreeCelsius}), indicating regional heterogeneity.

Analysis of multi-day lead errors (see Appendix) shows that the error magnitude does not increase substantially on the second day. These findings contradict the assumption of monotonically increasing forecast standard deviation with lead time, reported in~\cite{Antoniadou-Plytaria2022}.

\begin{table}[bt]
  \caption{HRRRv4 forecast accuracy at SOLRAD network stations and selected \ac{PV} systems with 24 leads at 6:00~UTC.}
  \label{tab:weather_forecast-all_stations}
  \footnotesize{}
  \centering{}
  \begin{tabular}{c c c c c c c}
    \toprule{}
    & \multicolumn{2}{c}{GHI in \SI[per-mode=symbol]{}{\watt\per\metre\squared}} & \multicolumn{2}{c}{Temperature in \SI{}{\degreeCelsius}}                \\
    Station        & MBE                                    & MAE                                               & MBE   & MAE  \\
    \midrule{}
    System 33      & 20.68                                  & 39.94                                             & 3.88  & 3.97 \\ \vspace{0.1cm}
    System 1423    & 12.66                                  & 22.74                                             & -1.02 & 1.61 \\
    Albuquerque    & 11.65                                  & 27.89                                             & 0.77  & 1.41 \\
    Bismarck       & 14.00                                  & 31.05                                             & 0.26  & 1.34 \\
    Hanford        & 6.04                                   & 20.17                                             & -1.88 & 2.10 \\
    Madison        & 10.45                                  & 35.30                                             & -0.29 & 1.18 \\
    Seattle        & 4.83                                   & 34.28                                             & 0.13  & 1.26 \\
    Salt Lake City & 9.48                                   & 28.87                                             & 1.57  & 1.91 \\
    Sterling       & 15.94                                  & 36.24                                             & -0.74 & 1.28 \\
    \midrule{}
    Average        & 11.75                                  & 30.72                                             & 0.30  & 1.78 \\
    \bottomrule{}
  \end{tabular}
\end{table}

The \ac{GHI} forecast error distribution is unimodal for all leads and is sharply peaked around zero, indicating frequent perfect or near-perfect predictions (cf. Figures~\ref{fig:wfm_33_lead_7}--\ref{fig:wfm_33_lead_13} and~\ref{fig:wfm_1423_lead_7}--\ref{fig:wfm_1423_lead_13}). A systematic bias is apparent: forecasts tend to slightly underestimate irradiance in the morning (leads~6--8) and to overestimate during later daytime hours. This temporal pattern is reflected in the skewness, which is strongly negative in the early leads (below \(-1.0\)) and strongly positive during the day (up to \(3.27\) for system~1423). The distributions are leptokurtic, with kurtosis values of at least \(5\) for both systems. Forecasts at the location of system~1423 are generally more accurate than at system~33, exhibiting a higher kurtosis (11.66 versus 5.23) and a lower \ac{MAE} (\SI[per-mode=symbol]{22.74}{\watt\per\metre\squared} versus \SI[per-mode=symbol]{39.94}{\watt\per\metre\squared}); see Table~\ref{tab:error_summary}.

\subsection{Combined Model}
The plant characteristic model is trained on satellite data and thus ignores the systematic \ac{GHI} bias in the \ac{NWP} forecasts. Training on forecasts might let it compensate for that bias, but the model is deliberately plant-specific and not a bias corrector; this modularity lets the \ac{NWP} part be updated separately. Bias can instead be handled by post-processing the \ac{NWP} output or by using an ensemble of \acp{NWP}~\cite{Lu2015}.

In the combined-model, the plant model is driven by daily weather forecasts initialized at 06:00~UTC rather than by ground truth. Error metrics computed on these combined forecasts quantify the influence of the \ac{GHI} bias: the negative bias observed for system~1423 in the plant-only evaluation is partly offset by the positive \ac{GHI} bias, yielding a slight positive \ac{MBE} of \SI[per-mode=symbol]{6.57}{\watt\per\kWp} (Table~\ref{tab:error_summary}). System~33 likewise exhibits the effect of the \ac{GHI} overestimation.

The \ac{MAE} of approximately \SI[per-mode=symbol]{28}{\watt\per\kWp} for both systems (plant-only) increases to \SI[per-mode=symbol]{31.16}{\watt\per\kWp} for system~1423 (an 11\% increase) and to \SI[per-mode=symbol]{47.28}{\watt\per\kWp} for system~33 (a 68\% increase). These results demonstrate that spatial differences in weather-forecast accuracy have a substantial and measurable impact on \ac{PV} power-forecast performance.

\begin{figure}[tb]
  \centering{}
  \begin{subfigure}{0.9\linewidth}
    \includegraphics[width=\linewidth]{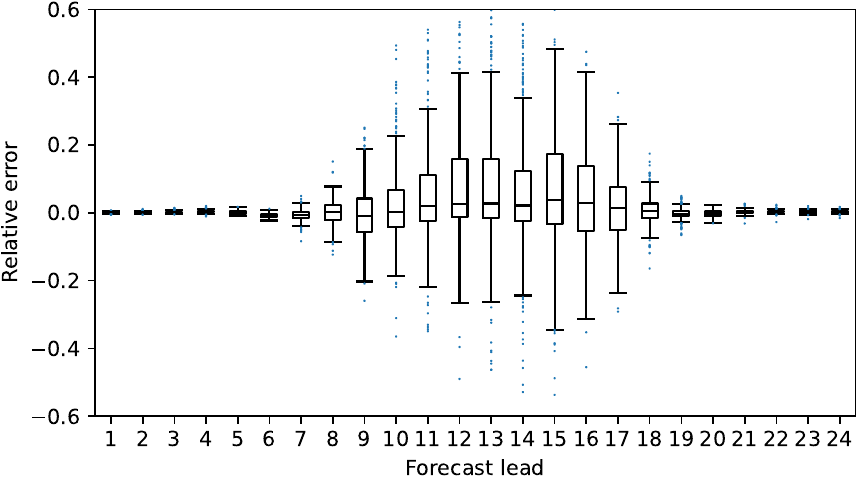}
    \caption{Relative error of different leads.}
    \label{fig:cm_33_boxplot}
    \vspace{0.2cm}
  \end{subfigure}
  \begin{subfigure}{0.45\linewidth}
    \includegraphics[width=\linewidth]{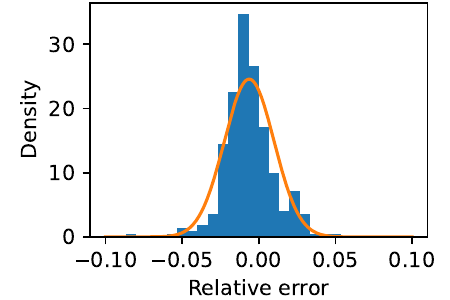}
    \caption{Distribution for lead 7}
    \label{fig:cm_33_lead_7}
    \vspace{0.1cm}
  \end{subfigure}
  \begin{subfigure}{0.45\linewidth}
    \includegraphics[width=\linewidth]{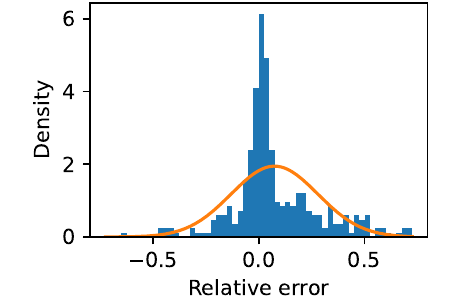}
    \caption{Distribution for lead 13}
    \label{fig:cm_33_lead_13}
    \vspace{0.1cm}
  \end{subfigure}
  \caption{Errors of the combined model with a forecast at 23:00 of system 33. Orange curves show the fitted normal distribution in the probability density plots.}
\end{figure}

\begin{figure}[tb]
  \centering{}
  \begin{subfigure}{0.9\linewidth}
    \includegraphics[width=\linewidth]{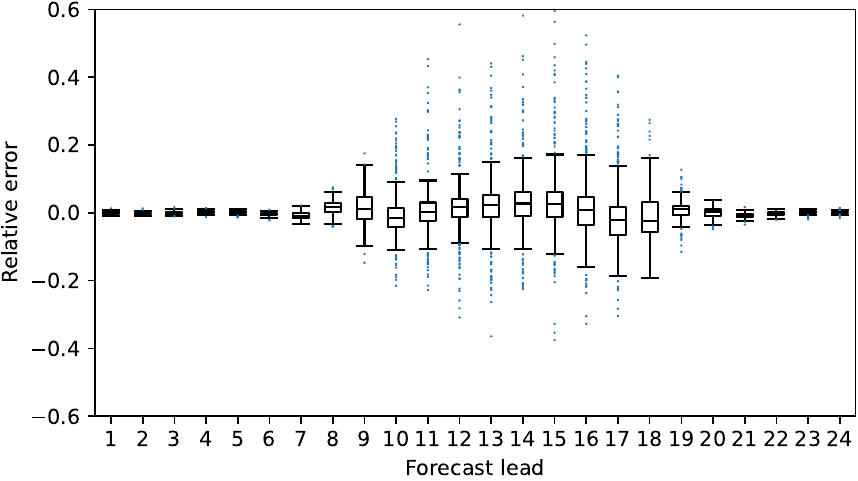}
    \caption{Relative error of different leads.}
    \label{fig:cm_1423_boxplot}
    \vspace{0.2cm}
  \end{subfigure}
  \begin{subfigure}{0.45\linewidth}
    \includegraphics[width=\linewidth]{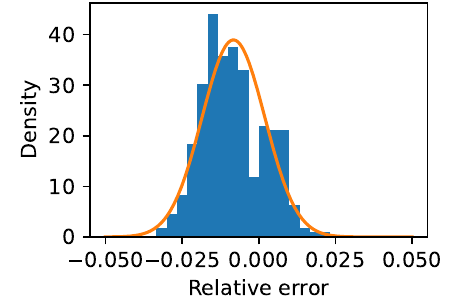}
    \caption{Distribution for lead 7}
    \label{fig:cm_1423_lead_7}
    \vspace{0.1cm}
  \end{subfigure}
  \begin{subfigure}{0.45\linewidth}
    \includegraphics[width=\linewidth]{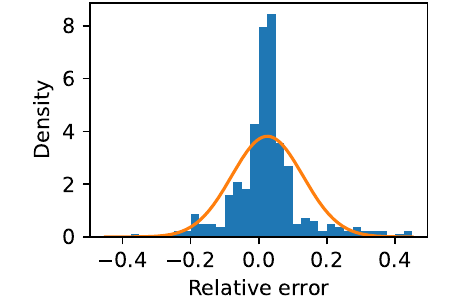}
    \caption{Distribution for lead 13}
    \label{fig:cm_1423_lead_13}
    \vspace{0.1cm}
  \end{subfigure}
  \caption{Errors of the combined model with a forecast at 22:00 of system 1423. Orange curves show the fitted normal distribution in the probability density plots.}
\end{figure}

Consistent with the \ac{GHI} analysis, the combined-model errors display a small negative bias during morning leads (visible in Figures~\ref{fig:cm_1423_boxplot} and~\ref{fig:cm_33_boxplot}). The skewness and kurtosis patterns of the combined-model error distributions mirror those of the \ac{GHI} forecasts but with slightly reduced intensity, as exemplified in Figures~\ref{fig:cm_33_lead_7}, \ref{fig:cm_33_lead_13}, \ref{fig:cm_1423_lead_7}, and~\ref{fig:cm_1423_lead_13}.

We fitted normal, generalized hyperbolic, and Student's t distributions to lead-wise errors and tested goodness-of-fit with Kolmogorov-Smirnov and Cramér-von Mises~\cite{csorgHo1996exact} tests (\( \alpha=0.05 \)) on diurnal leads (6--20). For system 1423 (Table~\ref{tab:distribution_fitness}) the normal is rejected for leads 10-19, while Student's t and generalized hyperbolic fit substantially better. Q--Q plots are in the Appendix.

\begin{figure}[bt]
  \centering{}
  \includegraphics[width=0.9\linewidth]{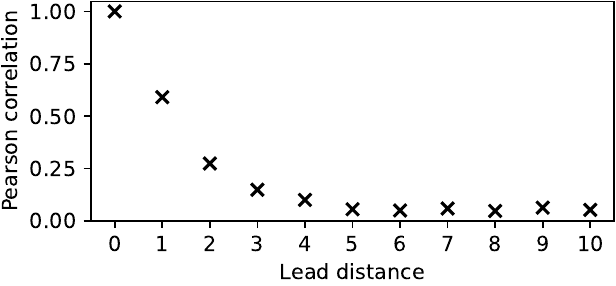}
  \caption{Temporal correlations of lead distance using SOLRAD stations locations forecasts for every 6 hours}
  \label{fig:temporal_correlation}
\end{figure}

Another aspect is the temporal dependence of forecast errors. If, for example, clear sky weather is predicted at 13:00, the forecast error at 14:00 will be small as well. To evaluate this assumption, we computed the Pearson correlation coefficient between the error at each forecast lead and errors at subsequent leads for forecasts initialized at 06:00~UTC\@. As shown in Figure~\ref{fig:temporal_correlation}, the lag-1 correlation is \( 0.58 \) and decays with increasing lead separation. Multi-period stochastic optimization models should therefore account for this temporal autocorrelation when representing forecast uncertainty.

\begin{table}[tb]
  \caption{Goodness-of-Fit of different distributions for system 1423 showing the p-values of the Kolmogorov-Smirnov (K-S) and Cramér-von Mises (C-M) test at 22:00 local time.}
  \label{tab:distribution_fitness}
  \scriptsize{}
  \centering{}
  \begin{tabular}{ c c c c c c c }
    \toprule{}
    & \multicolumn{2}{c}{Normal} & \multicolumn{2}{c}{Student's t} & \multicolumn{2}{c}{\shortstack{Generalized                         \\ Hyperbolic}} \\
    Lead & K-S                        & C-M                             & K-S                                        & C-M   & K-S   & C-M   \\
    \midrule%
    6    & 0.615                      & 0.445                           & 0.591                                      & 0.419 & 0.576 & 0.474 \\
    7    & 0.160                      & 0.092                           & 0.090                                      & 0.158 & -     & 0.009 \\
    8    & 0.735                      & 0.866                           & 0.735                                      & 0.866 & 0.597 & 0.811 \\
    9    & 0.634                      & 0.612                           & 0.590                                      & 0.596 & 0.864 & 0.845 \\
    10   & 0.000                      & 0.000                           & 0.049                                      & 0.104 & 0.534 & 0.511 \\
    11   & 0.000                      & 0.000                           & 0.434                                      & 0.692 & 0.163 & 0.394 \\
    12   & 0.000                      & 0.000                           & 0.587                                      & 0.541 & 0.237 & 0.418 \\
    13   & 0.000                      & 0.000                           & 0.504                                      & 0.580 & 0.570 & 0.642 \\
    14   & 0.000                      & 0.000                           & 0.648                                      & 0.578 & 0.294 & 0.595 \\
    15   & 0.000                      & 0.000                           & 0.428                                      & 0.499 & 0.553 & 0.446 \\
    16   & 0.000                      & 0.000                           & 0.290                                      & 0.366 & 0.323 & 0.566 \\
    17   & 0.000                      & 0.000                           & 0.076                                      & 0.085 & 0.414 & 0.639 \\
    18   & 0.008                      & 0.004                           & 0.107                                      & 0.049 & 0.992 & 0.971 \\
    19   & 0.001                      & 0.002                           & 0.166                                      & 0.133 & 0.224 & 0.178 \\
    20   & 0.520                      & 0.513                           & 0.600                                      & 0.528 & 0.475 & 0.432 \\
    \bottomrule{}
  \end{tabular}
\end{table}

\section{CONCLUSIONS}\label{sec:conclusions}
Knowledge of the forecast error distribution of renewable energy generation is a critical input for stochastic optimization. Motivated by a lack of reference distributions for \ac{PV} systems in the literature, we conducted a detailed analysis of forecast errors using a two-part model.
The plant-characteristic component is an ensemble of \acp{MLP} that captures plant-specific behavior, including panel size, orientation, efficiency, and local shading arising from the surrounding environment and weather. Its \ac{MAE} for two exemplary systems is approximately \SI[per-mode=symbol]{28}{\watt\per\kWp} when satellite-based observation data are treated as a perfect weather forecast. The second component is an \ac{NWP}-based weather forecast model, in particular the \ac{HRRR}v4 model, applied to the most relevant weather features (\ac{GHI} and air temperature). This model exhibits spatially varying accuracy and a systematic bias in \ac{GHI} forecasts, resulting in a modest overestimation. Combined, the two components increase the \ac{MAE} by 11\% and 68\% for the two examined \ac{PV} systems.

The common Gaussian assumption for forecast errors does not adequately describe the observed distributions. The generalized hyperbolic and Student's t distributions provide better fits. Additionally, forecast errors across different lead times exhibit temporal correlation, which decreases with increasing lead separation and thus should be considered in applications.

For future work, we recommend investigating the potential benefits of combining multiple \ac{NWP} models in an ensemble, as suggested by~\cite{Lu2015}, or introducing a statistical intermediate layer to mitigate the \ac{GHI} bias. Such measures could reduce forecast errors and improve the accuracy of \ac{PV} system forecasts. Further, a similar two-part modelling approach could be applied to wind and load forecasts to analyze the interplay of the two-parted model.

\section*{\uppercase{Acknowledgements}}
This work has received funding from the European Union's Horizon 2020 research and innovation programme under grant agreement No 957819. The authors want to thank Johannes Heil for implementation, feedback and discussions.
Microsoft Copilot has been used for language improvements using a handcrafted initial version of the script as input.

\bibliographystyle{apalike}
{
  \small{}
  \bibliography{bibliography}
}

\section*{\uppercase{Appendix}}
\begin{figure*}[tb]
  \centering{}
  \begin{subfigure}{0.482\linewidth}
    \includegraphics[width=\linewidth]{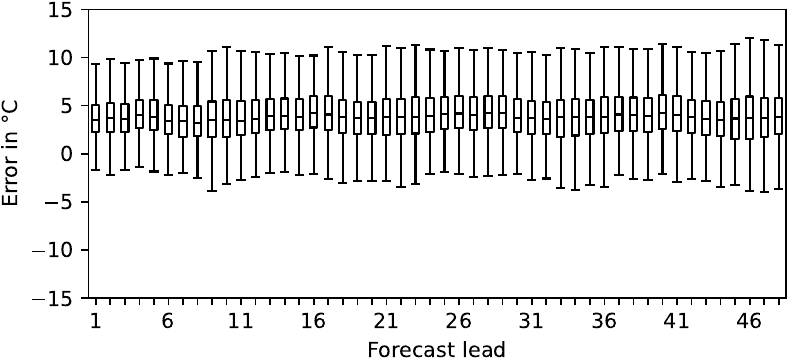}
    \caption{System 33.}
    \vspace{10pt}
  \end{subfigure}
  \hspace{10pt}
  \begin{subfigure}{0.482\linewidth}
    \includegraphics[width=\linewidth]{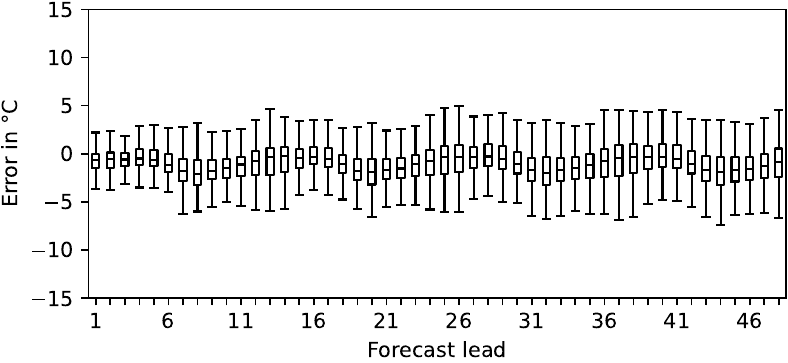}
    \caption{System 1423.}
    \vspace{10pt}
  \end{subfigure}
  \begin{subfigure}{0.24\linewidth}
    \includegraphics[width=\linewidth]{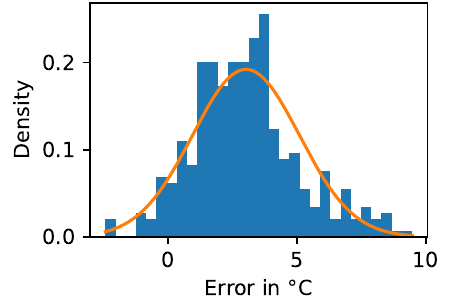}
    \caption{System 33 lead 12.}
  \end{subfigure}
  \begin{subfigure}{0.24\linewidth}
    \includegraphics[width=\linewidth]{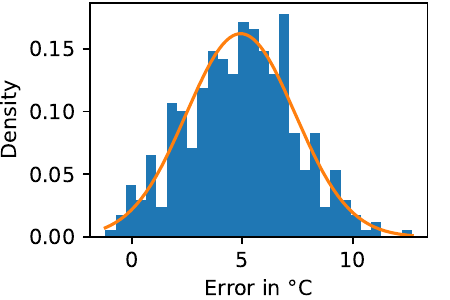}
    \caption{System 33 lead 24.}
  \end{subfigure}
  \begin{subfigure}{0.24\linewidth}
    \includegraphics[width=\linewidth]{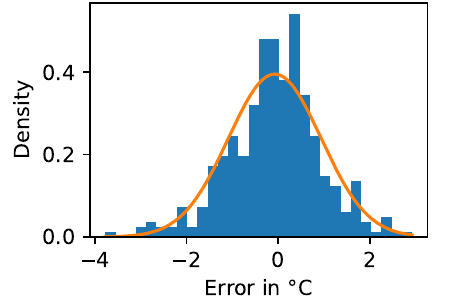}
    \caption{System 1423 lead 12}
  \end{subfigure}
  \begin{subfigure}{0.24\linewidth}
    \includegraphics[width=\linewidth]{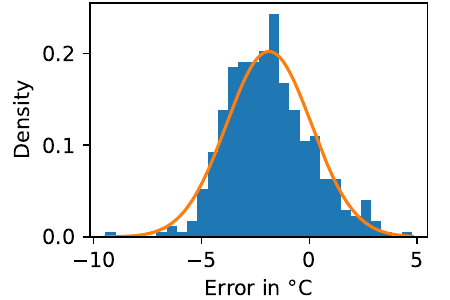}
    \caption{System 1423 lead 24}
  \end{subfigure}
  \vspace{10pt}
  \caption{Errors of the weather forecast model for air temperature with a forecast at 6:00 UTC.}
  \label{fig:wfm_33_temp}
\end{figure*}

\paragraph*{Weather forecast model temperature error}\label{apx:weather_forecast_temperature}
Compared to \ac{GHI}, air temperature forecast has an almost constant error distribution as shown in Figure~\ref{fig:wfm_33_temp}. Slight variations due to the day-night cycle can be observed. Further, a bias of around \( 4^\circ C \) is visible for system 33. The error is almost normally distributed for all leads.

\begin{figure*}[tb]
  \centering{}
  \begin{subfigure}{0.22\linewidth}
    \includegraphics[width=\linewidth]{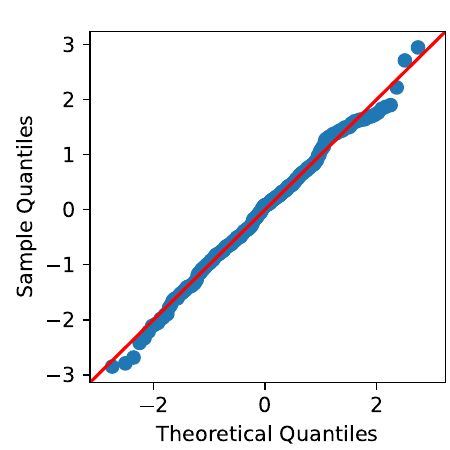}
    \caption{Normal, lead 8}
  \end{subfigure}
  \begin{subfigure}{0.22\linewidth}
    \includegraphics[width=\linewidth]{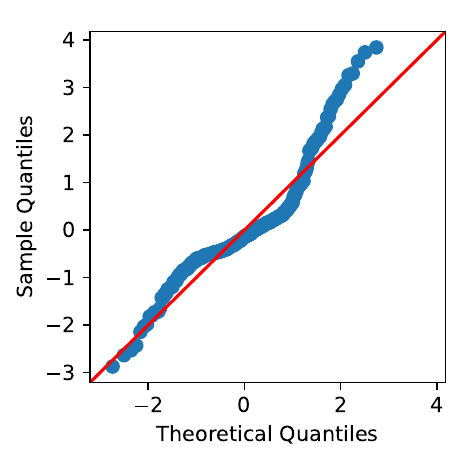}
    \caption{Normal, lead 10}
  \end{subfigure}
  \begin{subfigure}{0.22\linewidth}
    \includegraphics[width=\linewidth]{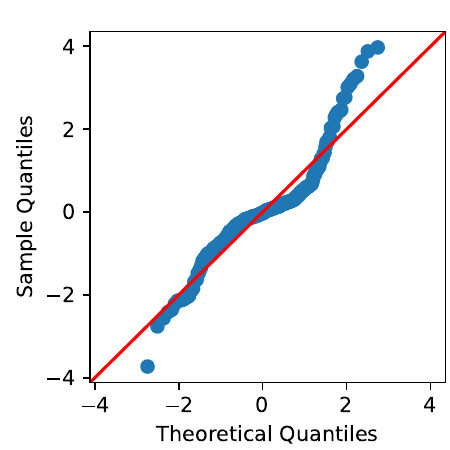}
    \caption{Normal, lead 13}
  \end{subfigure}
  \begin{subfigure}{0.22\linewidth}
    \includegraphics[width=\linewidth]{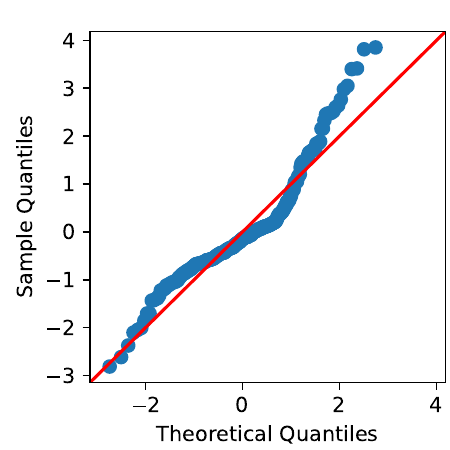}
    \caption{Normal, lead 17}
  \end{subfigure}

  \vspace{5pt}
  \begin{subfigure}{0.22\linewidth}
    \includegraphics[width=\linewidth]{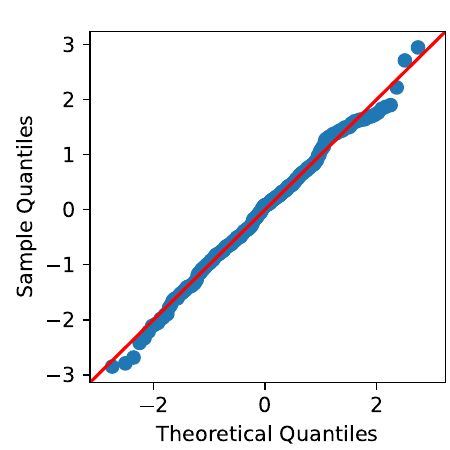}
    \caption{Student's t, lead 8}
  \end{subfigure}
  \begin{subfigure}{0.22\linewidth}
    \includegraphics[width=\linewidth]{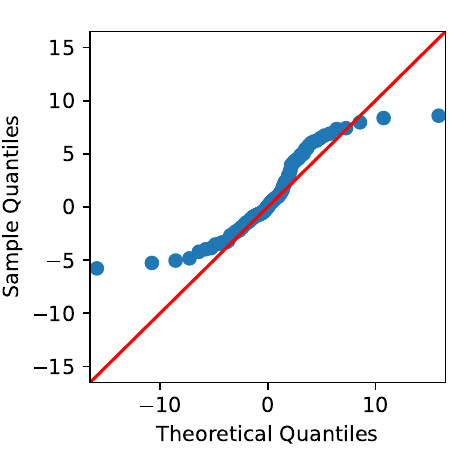}
    \caption{Student's t, lead 10}
  \end{subfigure}
  \begin{subfigure}{0.22\linewidth}
    \includegraphics[width=\linewidth]{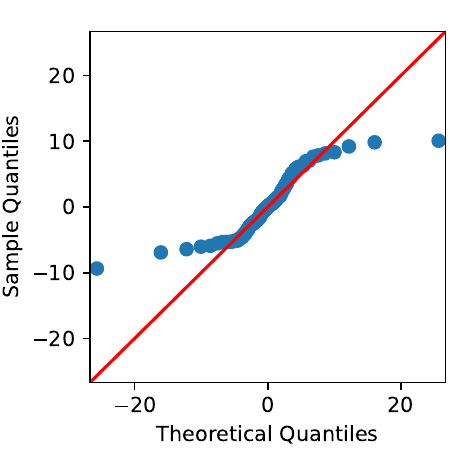}
    \caption{Student's t, lead 13}
  \end{subfigure}
  \begin{subfigure}{0.22\linewidth}
    \includegraphics[width=\linewidth]{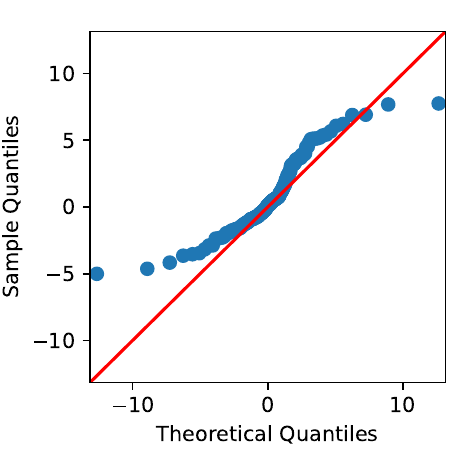}
    \caption{Student's t, lead 17}
  \end{subfigure}

  \vspace{5pt}
  \begin{subfigure}{0.22\linewidth}
    \includegraphics[width=\linewidth]{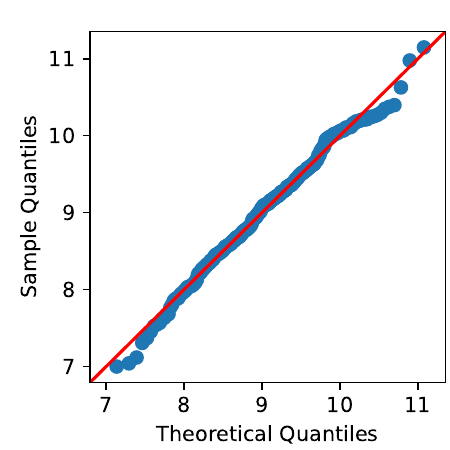}
    \caption{Hyperbolic, lead 8}
  \end{subfigure}
  \begin{subfigure}{0.22\linewidth}
    \includegraphics[width=\linewidth]{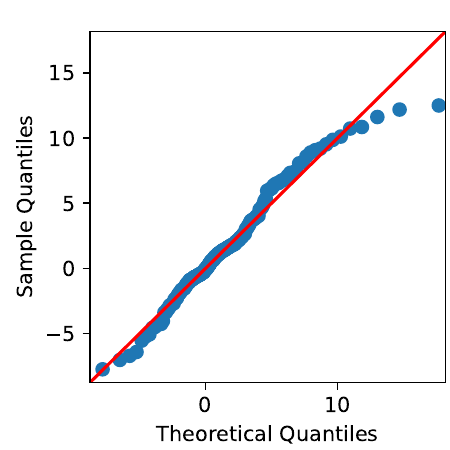}
    \caption{Hyperbolic, lead 10}
  \end{subfigure}
  \begin{subfigure}{0.22\linewidth}
    \includegraphics[width=\linewidth]{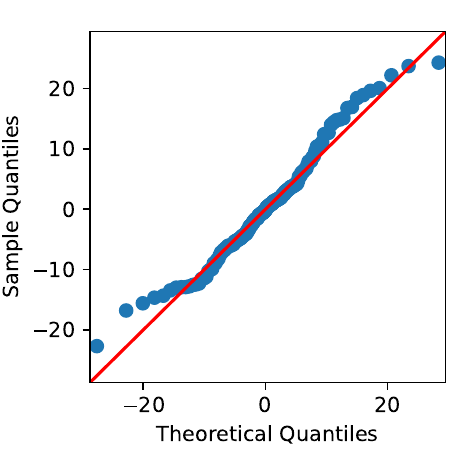}
    \caption{Hyperbolic, lead 13}
  \end{subfigure}
  \begin{subfigure}{0.22\linewidth}
    \includegraphics[width=\linewidth]{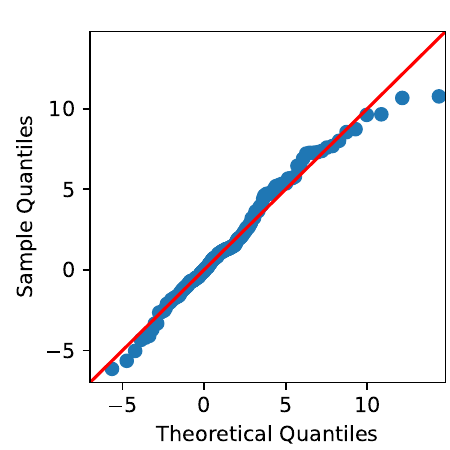}
    \caption{Hyperbolic, lead 17}
  \end{subfigure}
  \vspace{10pt}
  \caption{Q-Q plots on empirical data against fitted distributions for the forecast at 22:00 for system 1423.}
  \label{fig:qq-plots}
\end{figure*}

\paragraph*{Distribution Fitting} \label{apx:distribution_fit}
Q-Q plots for selected leads of system 1423 (Figure~\ref{fig:qq-plots}) show the normal distribution fits poorly(first row), Student's t fits the center but misses the tails (second row), and the generalized hyperbolic fits best overall (third row). All three perform well at morning lead 8, while generalized hyperbolic is superior for higher daytime leads.

\end{document}